\documentclass[conference]{IEEEtran}
\IEEEoverridecommandlockouts

\usepackage{hyperref}
\usepackage{pgfplots, tikz}
\pgfplotsset{compat=1.17}
\usepackage{cite}
\usepackage{amsmath,amssymb,amsfonts}
\usepackage{algorithmic}
\usepackage{graphicx}
\usepackage{colortbl}
\usepackage{float}
\usepackage{caption}
\usepackage{subcaption}

\usepackage{multirow}
\usepackage{multicol}
\usepackage{booktabs}
\usepackage{textcomp}
\usepackage{xcolor}
\def\BibTeX{{\rm B\kern-.05em{\sc i\kern-.025em b}\kern-.08em
    T\kern-.1667em\lower.7ex\hbox{E}\kern-.125emX}}
\begin{document}

\title{Hardware-aware Graph Neural Networks prunning for embedded event-based vision}

%\author{Authors are removed for blind review}
\author{
%\IEEEauthorblockN{Removed for blind review}
\IEEEauthorblockN{Piotr Wzorek, Kamil Jeziorek and Tomasz Kryjak\thanks{This work was supported by the “Excellence initiative – research university” programme for AGH University of Krakow, the Polish National Science Centre projects 2024/53/N/ST6/04254 and 2024/53/N/ST6/04331, Polish high-performance computing infrastructure PLGrid (HPC Center: ACK Cyfronet AGH – grant no. PLG/2025/017956) and 16.16.120.773 for the AGH University of Krakow.}}
\IEEEauthorblockA{\textit{Embedded Perception and Autonomous Systems Group, AGH University of Krakow, Poland} \\
\textit{pwzorek@agh.edu.pl, 	kjeziorek@agh.edu.pl, tomasz.kryjak@agh.edu.pl}} 
}

\maketitle

\begin{abstract}

Event-based cameras are gaining popularity as the sensor of choice for mobile robotics, due to their high performance in dynamic environments.  
However, these applications require efficient real-time data processing with low latency and power consumption. 
One strategy to meet these stringent requirements is hardware acceleration of efficient algorithms that preserve the temporal sparsity of event data.  

In this work, we propose an optimization strategy for Graph Convolutional Neural Networks models aimed at adapting their architecture to the limited resources of embedded heterogeneous FPGA platforms. 
Our method incorporates hardware-aware pruning and quantization, taking into account the trade-off between on-chip memory savings and inference accuracy.  
Strategic exploration of the design space with Fine Grid Search and Greedy layer-wise Iterative Deepening Search methods enables flexible adaptation of the model architecture to the target platform.
Our approach was evaluated across various network configurations and multiple datasets, resulting in BRAM memory reductions of 28.8\% for CIFAR-10 (with a 1.65\% decrease in accuracy), 31.4\% for MNIST-DVS (accuracy drop of 3.55\%), and 26.5\% for N-Caltech101 (with a 5.18\% accuracy reduction).

\end{abstract}


\section{Introduction}
\label{sec:intro}

Vision systems in mobile robotics are subject to stringent requirements in terms of prediction accuracy, latency, throughput, and energy efficiency. 
High system performance is also essential under challenging conditions—such as high-speed motion or adverse lighting environments.
Meeting these demands necessitates a high-dynamic-range sensor, an efficient algorithm, and a specialised hardware platform.

Event cameras (also known as Dynamic Vision Sensors - DVS) are increasingly being used as an alternative to conventional vision sensors.
These sensors register changes in the observed scene independently for each pixel \cite{gallego2020event}.
The data captured in this manner form a sparse point cloud in the space-time, where each point is characterised by four values: a timestamp (with a resolution of microseconds), pixel coordinates, and polarity (indicating the direction of brightness change — positive or negative).
With their operating principle, event cameras are distinguished by low latency, low energy consumption, and a high dynamic range.

Efficient processing of event-based data poses a significant scientific challenge.
Early approaches employed two- or three-dimensional representations obtained through data accumulation and subsequently used as input to Convolutional Neural Networks \cite{perot2020learning} or Vision Transformers \cite{gehrig2023recurrent}.
However, such representations tend to compromise the sparsity of data, resulting in a substantial amount of redundant computation.

The described issue has been addressed through the use of Spiking Neural Networks (SNN) \cite{gehrig2020event}.
However, such systems often lack the accuracy achieved by alternative learning methods.
Moreover, their efficient hardware acceleration on platforms (other than neuromorphic) is challenging, due to the non-deterministic pattern of synaptic weight memory access.

Consequently, a commonly considered strategy nowadays involves the use of Graph Convolutional Neural Networks (GCNNs), where each event is treated as a node in a graph, with edges defining local dependencies between them \cite{schaefer2022aegnn}\cite{gehrig2024low}. This method demonstrates high accuracy and, moreover, is well-suited for hardware acceleration.

In the context of mobile robotics, compact models that can be deployed on embedded low-power systems (enabling autonomous operation, e.g. onboard a drone) are highly desirable.
Based on these assumptions, we propose hardware-aware pruning and quantization specifically designed for the EFGCN (Event-based FPGA-accelerated Graph Convolutional Network) architecture \cite{jeziorek2024embedded}.
In our work, we primarily focus on reducing the utilisation of block memory resources, which often constitute the main bottleneck in the hardware implementation of larger models.
This is particularly important when using external DRAM is not possible or advisable due to energy or latency constraints.
Our method involves selecting a model from the design space to ensure an appropriate trade-off between inference accuracy and internal memory usage of the FPGA. The main contributions of our paper can be summarised as follows:

\begin{itemize}
    \item We present the first hardware-aware pruning and quantization method for event-based vision and the first such strategy for Graph Convolutional Neural Networks.
    \item The proposed method enables flexible selection of the final solution depending on the requirements of the specific application and the target platform.
    \item We have evaluated the method for various datasets and architectures achieving between 26.5\% - 31.4\% reduction of the number of BRAMs utilised for feature vectors with an accuracy drop between 1.65\% and 5.18\%, 
    \item We have developed a hardware proof-of-concept module demonstrating the effectiveness of the proposed method.
\end{itemize}

The remainder of this paper is organised as follows.
In Section~\ref{sec:related}, we introduce existing works related to our study while Section~\ref{sec:method} describes the method proposed in this work. In Section~\ref{sec:results}, we present the results of the conducted experiments. The paper concludes with a discussion of the proposed solution in Section~\ref{sec:discussion}, followed by a summary in Section~\ref{sec:summary}.

\section{Related Work}
\label{sec:related}

\subsection{Event-based vision for FPGA}

While event-based vision systems have been actively developed for over a decade~\cite{chakravarthi2025recent}, the topic of their acceleration on FPGA platforms remains an emerging research area~\cite{kryjak2024event}.

To the best of our knowledge, only three systems designed for FPGAs have been proposed that enable real-time object classification based on event data.
The first of these~\cite{gao2024composable} is a system based on Sparse Convolutional Neural Networks. Despite its high prediction accuracy, the system exhibits significant FPGA resource utilisation and is intended for platforms that are not suitable for mobile robotics applications.

The remaining hardware solutions employ Graph Convolutional Neural Networks. In~\cite{yang2024evgnn}, a set of hardware modules was proposed that, by utilising directed graphs and \textit{PointNetConv}, enables fully asynchronous processing of event-based data with low latency. A limitation of this approach is its restricted scalability due to the absence of pooling layers — for higher data resolutions, the system demands substantial memory resources (including external memory, which results in non-deterministic latency and higher energy usage).

In this work, we have chosen to utilise the third system — \textbf{EFGCN}. In~\cite{jeziorek2024embedded}, the authors propose an architecture divided into two parts: in the first, data is processed asynchronously by the initial graph convolution layers. Subsequently, a 3D MaxPool layer is applied, which disrupts the temporal order of events (due to their accumulation within specified time windows), but reduces the amount of data that must be stored in memory. Furthermore, as a result, the required bandwidth of hardware modules can be determined in the synchronous part of the system, allowing for some computations to be performed sequentially, which leads to a reduction in logical resource usage. 
The system achieves high accuracy across most popular event-based datasets, providing high throughput and scalability.
In the present work, we address the optimisation of this solution.

\subsection{Pruning and quantisation}

In hardware-accelerated neural network models (including GCNNs for event-based vision) a significant portion of the internal FPGA memory is utilised for storing feature maps between successive layers. 
Reducing their size enables the implementation of larger models or the deployment on smaller, embedded platforms. 
Therefore, in this work, we focus on methods aimed at reducing the dimensionality of features -- \textbf{pruning} and \textbf{quantisation}.

Pruning involves the removal of a portion of parameters from the neural network model. Numerous methods are currently employed across a wide range of architectures — including graph neural networks.
The most popular pruning methods are magnitude-based, where weights with the smallest norms (e.g., L2 norm) are removed — their application is feasible for most types of networks and integrated into popular frameworks such as PyTorch.
However, the literature also presents more sophisticated strategies, some specifically proposed for graph neural networks. 

In the work \cite{wang2022pruning}, edge properties in a graph model were utilised to sparsify the processed graph, achieving a significant reduction in the number of floating-point operations per second (over 90\%) without a significant drop in accuracy. 
Conversely, \cite{huang2023cp3} addresses pruning for Point-based Networks (which serve as an inspiration for the \textit{PointNetConv} graph convolution employed in this work). The authors applied a coordinate-enhanced channel importance metric for pruning, which considers correlations between feature values within a network channel and the dimensional information of a specific point in the point cloud, achieving state-of-the-art performance for a sparse model.

However, among the numerous strategies for the selection of parameters to remove \cite{cheng2024survey}, in the context of the issue at hand, the main consideration should be given to methods that directly take into account the characteristics of the target platform.
For this purpose, the so-called structured pruning is used, where the selection of removed parameters is also determined by their position in the weight matrix. 
In the work of \cite{ramhorst2023fpga}, the known widths of the memory blocks were used to select the appropriate structure of the sets of parameters that are removed for a given model. 
In this way, the authors achieved a significant reduction in DSP and BRAM resources with minimal loss of prediction performance. 
Similarly, in \cite{plochaet2023hardware}, knowledge of the hardware architecture and the construction of computational elements were used to select the appropriate model from the design space. The solution thus achieved is not only sparser than the original one, but also adapted to the target hardware platform.

In this study we propose a hardware-aware pruning and quantisation approach, developed through an in-depth analysis of the EGFCN hardware module and the structural characteristics of memory elements in reconfigurable architectures.

\section{The proposed method}
\label{sec:method}

\begin{figure*}[!t]
    \centering
    \includegraphics[width=0.9\linewidth]{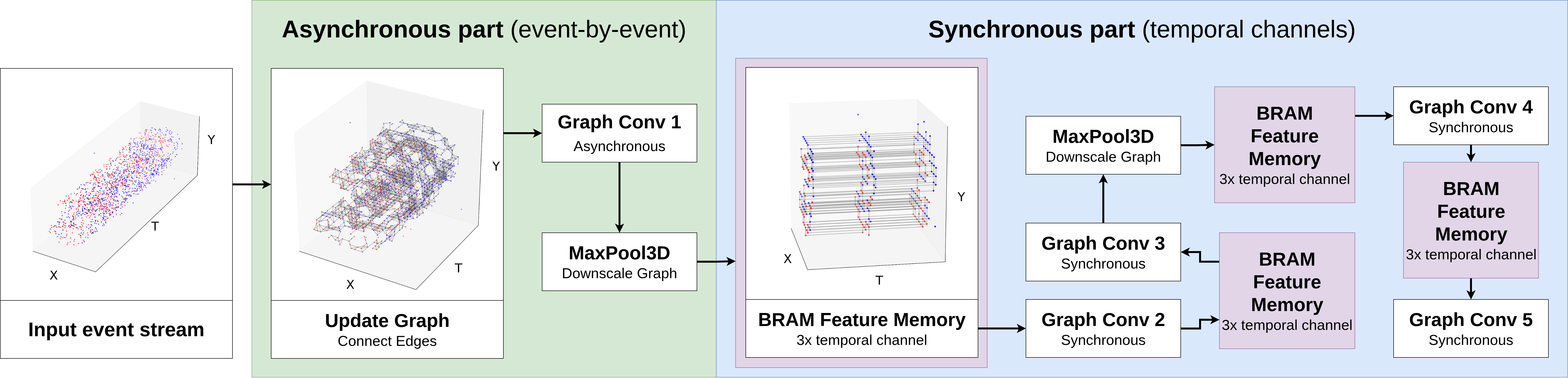}
    \caption{The overview of EFGCN architecture with highlighted asynchronous (green) and synchronous (blue) parts. Feature maps between successive layers in the model are stored in BRAM (purple).}
    \label{fig:overview}
\end{figure*}

%The method proposed in this work was developed based on an analysis of the hardware architecture of EFGCN, as introduced in \cite{jeziorek2024embedded}. 
The objective of our network pruning and quantisation method is to minimise the utilisation of internal FPGA memory resources (dual port 36kb BlockRAM and 288kb UltraRAM), and to enable their efficient usage.
Concurrently, we aim to preserve a high level of predictive accuracy in object classification tasks using event-based data.

\subsection{Preliminaries}

In the synchronous part of the EFGCN \cite{jeziorek2024embedded} architecture, convolutional layers and 3D MaxPooling layers are employed. 
The graph convolution \textit{PointNetConv} involves processing the feature vector associated with the currently considered vertex using a Multi-Layer Perceptron (MLP), followed by processing the feature vectors of all vertices connected to it (edges are defined in three dimensions — $x$, $y$, and time).

Between the layers, data are stored in so-called \textit{temporal channel} matrices — sub-graph representing events captured within a given time window (see Figure \ref{fig:overview}). 
The size of these subgraphs changes throughout the model, as it depends on the MaxPool layers, which lead to a rescaling of the representation.
These data are stored in a dual-port \textit{feature memory} implemented using either BlockRAM or UltraRAM (depending on the layer).
The required memory width depends on the number of elements in the feature vector (i.e. the number of channels $C$) and the precision of the quantised values $q$. 
Additionally, each memory element stores 18 bits of $edge$ information (describing the position of the connected vertices).
The memory depth depends on the size of the graph at a given layer.
Before each convolution layer we implement three independent \textit{feature memories} that store data for separate time windows (which is necessary as the vertices from different \textit{temporal channels} can be connected).

\subsection{BRAM and URAM utilisation}
\label{sec:bram_uram}

The number of BRAM utilised for the implementation of single \textit{feature memory} depends on the graph size (which is not subject to optimisation in this work) and the memory width:

\begin{align}
N_{18\mathrm{k}} &= \left\lceil \frac{C \cdot q + 18}{18} \right\rceil, \label{eq:halfbram}\\[4pt]
N_{\mathrm{BRAM}} &= \frac{N_{18\mathrm{k}}}{2}. \label{eq:bram}
\end{align}

The number of utilised URAM is given by the following formula:
\begin{equation}
N_{\mathrm{URAM}} = \left\lceil \frac{C \cdot q + 18}{72} \right\rceil. \label{eq:uram}
\end{equation}

To ensure full utilisation of memory resources, the number of channels $C$ for a layer quantised to $q$ bits must be chosen such that $N_{18\mathrm{k}}, N_{\mathrm{URAM}} \in \mathbb{N}  > 0$. 
This constraint leads to the following families of valid channel counts:

\begin{align}
C^{\text{(BRAM)}}_{m} &= m \cdot \frac{18}{g_{18}},         & m &\in\mathbb{Z}_{\ge 1},\label{eq:c_bram}\\[6pt] 
C^{\text{(URAM)}}_{n} &= C_{0} + n \cdot \frac{72}{g_{72}}, & n &\in\mathbb{Z}_{\ge 0}, \label{eq:c_uram}
\end{align}

where $g_{18} = \gcd(q, 18)$, $g_{72} = \gcd(q, 72)$, and $C_{0}$ is the smallest non-negative solution of the $C_{0}\,q \equiv 54 \pmod{72}$ congruence.
Selecting $C$ from the first family results in full utilisation of BRAM-18 kb halves (and BRAM-36 kb blocks for even $m$), whereas choosing $C$ from the second family ensures full utilisation of URAM-72 kb blocks.

\subsection{Searching method}

The goal of the search procedure is to find a configuration that uses the least hardware resources while maintaining high classification performance.
As indicated in Section~\ref{sec:bram_uram}, this can be achieved by pruning the channels $C$ and applying feature quantisation in accordance with equations~\eqref{eq:c_bram} and~\eqref{eq:c_uram}.
Assuming that the number of channels $C$ for each of the five convolutional layers in the model does not exceed the initial value described in Section~\ref{sec:results}, and considering two levels of quantisation, the maximum number of possible configurations exceeds $7.8 \cdot 10^6$.
Each configuration is assigned a utilisation cost, defined as the sum of BRAM resources used by each convolution. For the purpose of metric normalisation, one URAM block is treated as equivalent to two BRAM blocks. Due to the large configuration space, a two-stage search process is proposed, enabling the identification of a high-quality configuration at an acceptable computational cost. This process comprises the Fine Grid Search (FG) and Greedy Layer-wise Iterative Deepening (GLID) stages.

\subsubsection{Fine Grid Search (FG)}

This stage begins with a baseline configuration characterised by the maximum number of quantisation bits and no pruning.This configuration is expected to provide the best possible classification accuracy, but also the highest resource utilisation.
Starting from this configuration, $k$ downward steps are performed with respect to the number of output channels (pruning) for each supported quantisation level.
%, $k$ options are selected that satisfy the constraints of either \eqref{eq:c_bram} or \eqref{eq:c_uram}, depending on the type of memory used. 
The final search space is limited to a maximum size of:

\begin{align}
N_{conf} = (k \cdot b) ^ l 
\label{eq:num_conf}
\end{align}

where $b$ denotes the number of quantisation levels, and $l$ is the number of layers subjected to pruning. 
For each configuration within this search space, a classification score is computed. 
The best configuration is then selected based on the Pareto frontier, using a predefined trade-off criterion between classification quality and resource utilisation.

\subsubsection{Greedy layer-wise Iterative Deepening Search (GLID)}

At this stage, configurations are initially sorted in descending order according to BRAM utilisation separately for each layer. 
As a result, we obtain $l$ configuration sequences, with the initial configuration derived using the FG method. 
Subsequently, we iterate over each layer individually and apply a configuration that reduces utilisation by one step (with respect to either~\eqref{eq:c_bram} or~\eqref{eq:c_uram}). 
We select the modification of the layer that yields the best quality score, and in the next iteration, the search continues from this new configuration. 
In cases where more than one configuration with the same utilisation exists for a given layer, all such configurations are explored, and the best one is chosen. 
Consequently, at each step—which requires only $l$ searches—we ensure a gradual reduction of the overall model utilisation. 
The termination condition for this method is a specified maximum allowable difference in quality score between consecutive models. 
After determining the final configuration, the model undergoes fine-tuning to improve adaptation to the applied changes.

\section{Experimental results}
\label{sec:results}

\subsection{Setup}
\label{sec:sec:setup}

In our experiments, we utilised three popular neuromorphic datasets for the classification task. 
The MNIST-DVS \cite{mnistdvs} and CIFAR10-DVS \cite{cifardvs} datasets were recorded at a resolution of $128 \times 128$ pixels and consist of 10 classes each. 
The third dataset, N-Caltech101 \cite{ncaltech}, was recorded at a resolution of $240 \times 180$ pixels and comprises a total of 101 classes.

For our study, we utilised the $large$ model from \cite{jeziorek2024embedded}, which consists of five convolutional layers and two pooling layers. 
Initially, the model featured output channel counts of 16, 32, 64, 64, and 128 for the respective convolutional layers. 
However, these parameters did not satisfy the conditions specified in Section~\ref{sec:bram_uram}. 
Therefore, we changed the number of output channels in the model according to the formulas \eqref{eq:c_bram} and \eqref{eq:c_uram}. 
The initial configuration along with the utilised resources is summarised in Table~\ref{tab:start-config}. 
As the classifier, a double fully connected layer was employed, which was quantised to 8 bits and was not subjected to pruning.

The initial models were trained for 100 epochs using the Adam optimiser with an initial learning rate of $10 ^{-3}$, decaying to $10 ^ {-4}$ after 20 epochs, and a weight decay of $10^{-4}$. 
A combination of data augmentations was applied, including horizontal flipping, event rotation in the XY plane, and dropout between the two linear layers.

\begin{table}[!t]
\centering
\caption{Initial model configurations and the corresponding resource utilisation depending on the dataset.}
\resizebox{0.47\textwidth}{!}{%
\begin{tabular}{@{}lcccc@{}}
\toprule
       & \multicolumn{2}{c}{MNIST-DVS / Cifar10} & \multicolumn{2}{c}{N-Caltech101} \\ \midrule
Layer  & Memory Type        & Out Channels       & Memory Type    & Out Channels    \\ \midrule
Conv 1 & BRAM               & 18                 & URAM           & 24              \\
Conv 2 & BRAM               & 36                 & URAM           & 33              \\
Conv 3 & BRAM               & 63                 & BRAM           & 72              \\
Conv 4 & BRAM               & 72                 & BRAM           & 72              \\
Conv 5 & BRAM               & 135                & BRAM           & 135             \\ \bottomrule
\end{tabular}%
}

\label{tab:start-config}
\end{table}

\begin{figure*}[!t]
    \centering
    \begin{subfigure}[b]{0.32\textwidth}
        \centering
        \includegraphics[width=\linewidth]{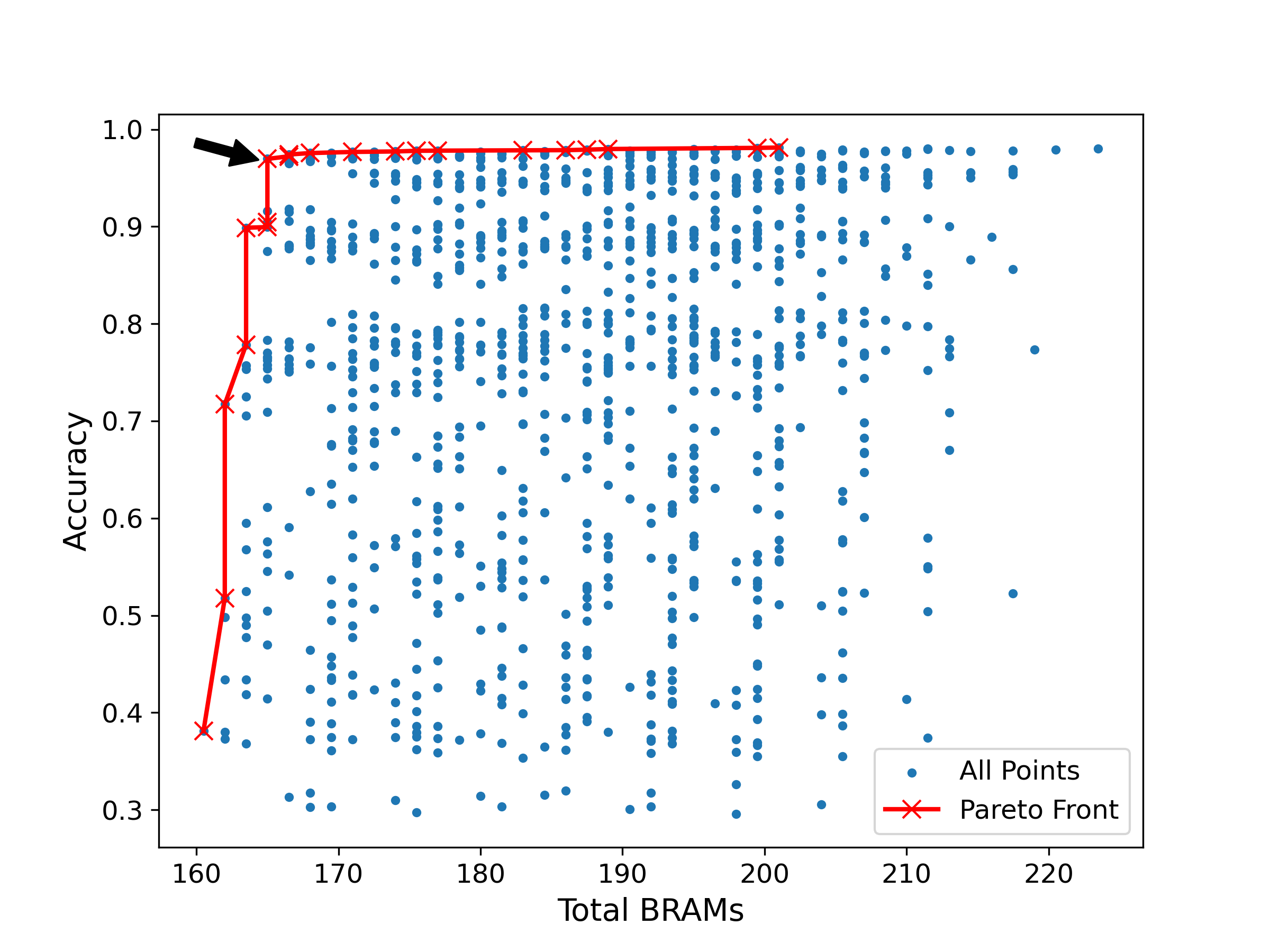}
        \caption{MNIST-DVS}
        \label{fig:mnist}
    \end{subfigure}
    \hfill
    \begin{subfigure}[b]{0.32\textwidth}
        \centering
        \includegraphics[width=\linewidth]{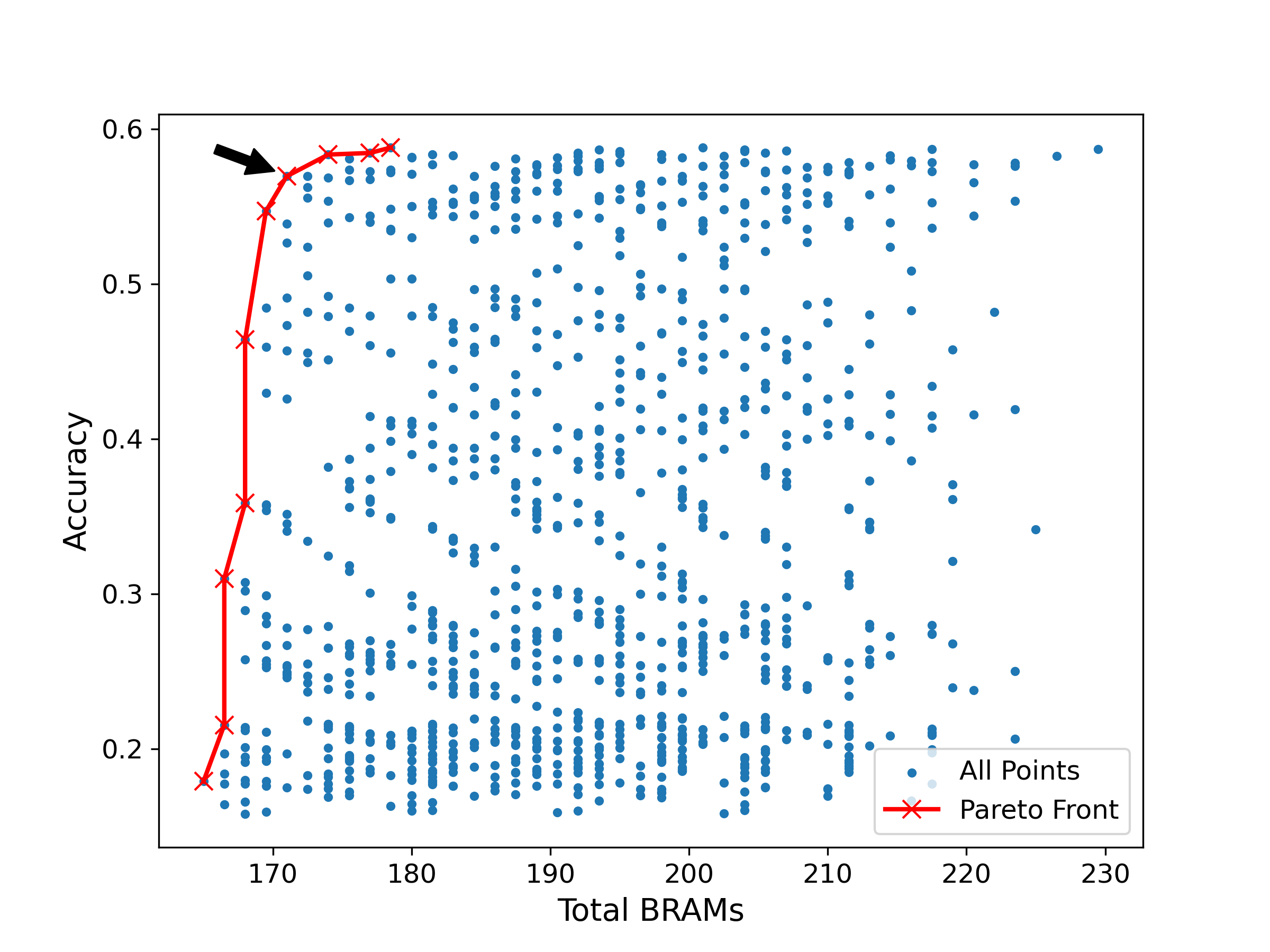}
        \caption{Cifar10-DVS}
        \label{fig:cifar}
    \end{subfigure}
    \hfill
    \begin{subfigure}[b]{0.32\textwidth}
        \centering
        \includegraphics[width=\linewidth]{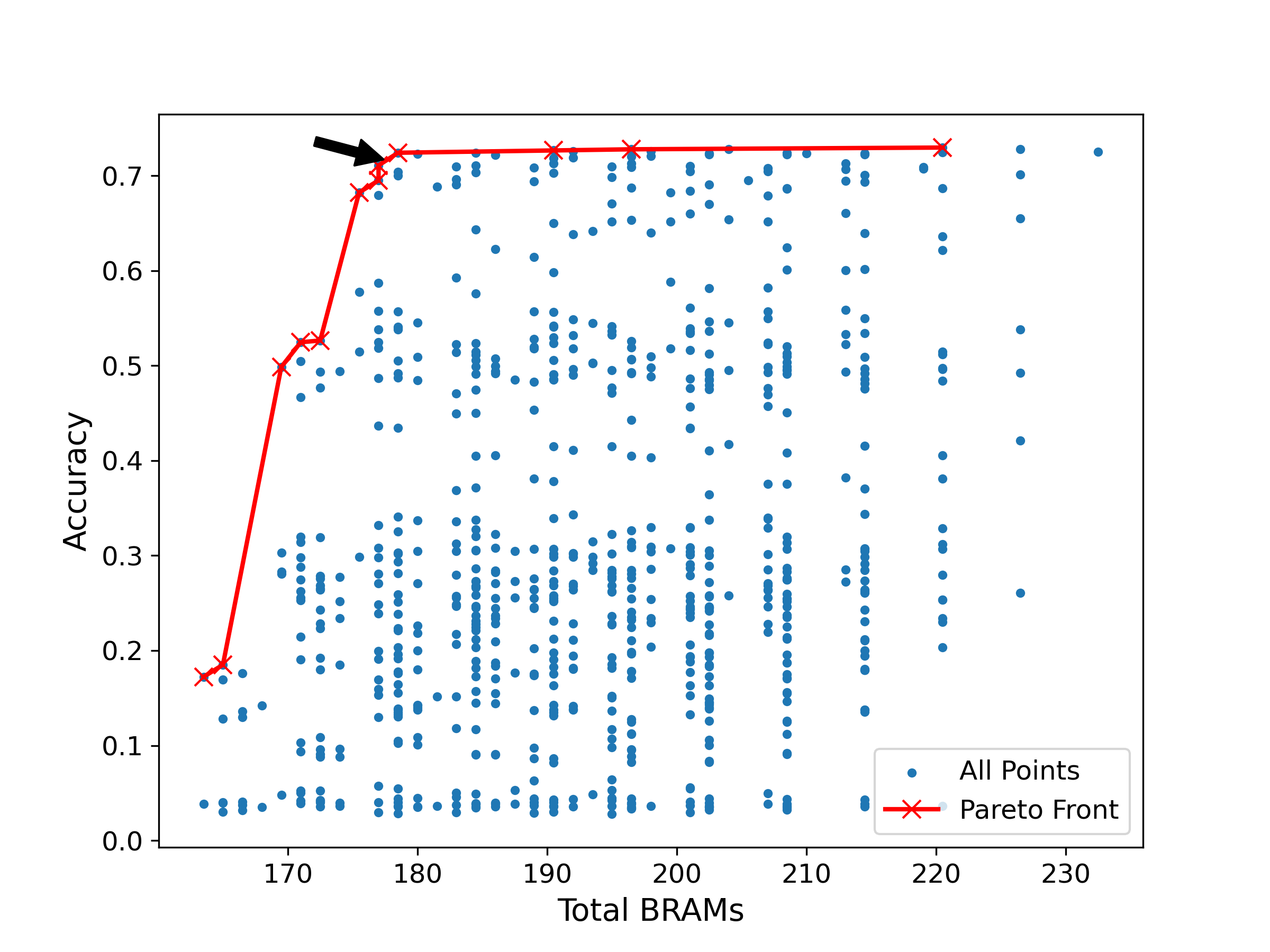}
        \caption{N-Caltech101}
        \label{fig:ncaltech101}
    \end{subfigure}
    \caption{Comparison of BRAM utilisation and classification performance for the MNIST-DVS, CIFAR10-DVS, and N-Caltech101 datasets, including the Pareto front with the selected configuration highlighted.}
    \label{fig:first_stage}
\end{figure*}

\subsection{First stage searching}
\label{sec:first_stage}

In our study, we selected two activation quantisation levels: 6 and 8 bits. 
We first established baseline accuracy and BRAM utilisation values for each model using 8-bit quantisation. 
These results, along with the initial floating-point model, are presented in Figure~\ref{fig:all-stages}. 
Subsequently, we determined a subset of configurations, applying $k=2$ steps for all datasets.
According to equation \eqref{eq:num_conf}, this corresponded to 1024 configurations. 
The identified search spaces together with the Pareto fronts are shown in Figure~\ref{fig:first_stage}.

The baseline models, visible as the rightmost points, achieve utilisation levels of approximately 229.5 and 232.5 BRAM blocks. 
Based on the derived Pareto frontier and its knee point—where the configurations marked by a black arrow were also selected—it is possible to significantly reduce BRAM utilisation by 55-58 blocks, with only a minor classification accuracy drop of 2\% for CIFAR10-DVS, 1\% for MNIST-DVS and 3.5\% for N-Caltech101. 
It is worth noting that these results are achieved without additional fine-tuning of the models.

At this stage, more stringent criteria can be established; for instance, knowing the resource constraints on a given platform, configurations with correspondingly lower RAM block usage can be selected.

\subsection{Second stage searching}

In the subsequent step, the configurations identified in Section~\ref{sec:first_stage} were further reduced in terms of BRAM utilisation using the GLID method. 
For the CIFAR10-DVS and N-Caltech101 datasets, the stopping criterion was set to a 5\% accuracy difference, while for MNIST-DVS it was set to 10\%. 
The final model configurations are presented in Table~\ref{tab:best-config}, and the utilisation and accuracy results are summarised in Figure~\ref{fig:all-stages}.

The stopping criterion was reached after 5, 6, and 4 steps for the CIFAR10-DVS, MNIST-DVS, and N-Caltech101 datasets, respectively. 
As a result, a reduction of approximately 7.5, 13.5, and 6 blocks was observed in Block RAM utilization, respectively.
The gains at this stage are not as significant compared to the FG method, and the quality metrics have decreased substantially relative to the baseline model, particularly for the MNIST model (in accordance with the stopping criterion). 
It is worth noting, however, that the configurations ultimately reached by this method are equivalent to those obtained via the FG method with $k=5$ steps, i.e. $10^5$ configurations — 97 times more than for $k=3$. 
An improvement to this method could involve modifying the search to allow backtracking to previous configurations in order to find better solutions.

\subsection{Finetuning}

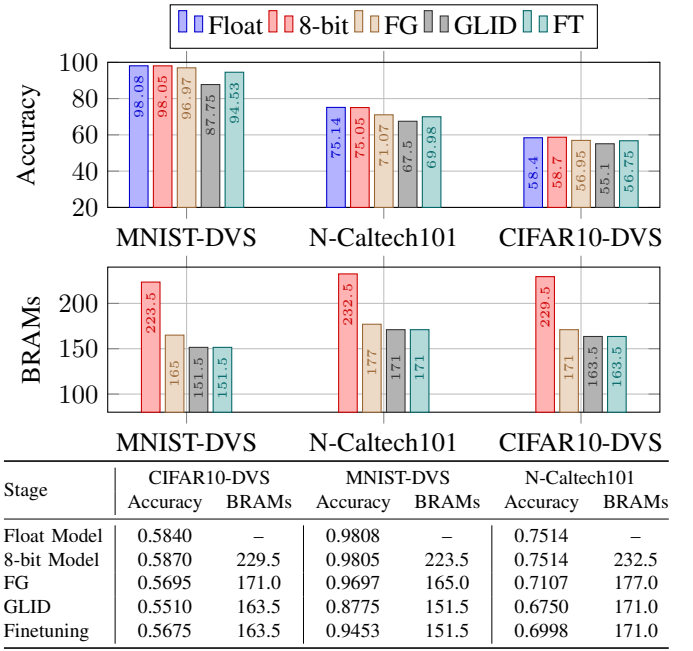
\begin{figure}[!t]
  \centering
  %===========================================
  % Podrys 1: tabela
  %===========================================
    \begin{subfigure}{0.49\textwidth}
    \centering
    \begin{tikzpicture}
    \begin{axis}[
            ylabel=Accuracy,
            name=plot1,
            symbolic x coords={MNIST-DVS,N-Caltech101,CIFAR10-DVS},
            height=3.5cm,
            width=\linewidth,
            legend style={at={(0.5,1.4)}, anchor=north, legend columns=-1},
            grid=major,
            ybar,
            xtick=data, 
            enlarge x limits=0.2,
            ymin=20, ymax=100,
            bar width=7pt,
            nodes near coords={
              \pgfmathprintnumber[
                 fixed zerofill,
                 precision=2
              ]{\pgfplotspointmeta}
            },
            nodes near coords align={vertical}, 
            every node near coord/.append style={
              font=\tiny,
              rotate=90,
              anchor=west,
              xshift=-23pt
            },
            ]
    \addplot+[
        draw=blue!80!black,
        fill=blue!30,
        nodes near coords,
        every node near coord/.append style={
            text=blue!80!black,
        }]
        coordinates {   (MNIST-DVS,98.08) 
                        (N-Caltech101,75.14)
                        (CIFAR10-DVS,58.40)
                        };               
    \addplot+[
        draw=red!80!black,
        fill=red!30,
        nodes near coords,
        every node near coord/.append style={
            text=red!80!black,
        }]
        coordinates {   (MNIST-DVS,98.05) 
                        (N-Caltech101,75.05)
                        (CIFAR10-DVS,58.70)
                        };                   
    \addplot+[
        draw=brown!80!black,
        fill=brown!30,
        nodes near coords,
        every node near coord/.append style={
            text=brown!80!black,
        }]
        coordinates {   (MNIST-DVS,96.97) 
                        (N-Caltech101,71.07)
                        (CIFAR10-DVS,56.95)
                        };                
    \addplot+[
        draw=black!80,
        fill=black!30,
        nodes near coords,
        every node near coord/.append style={
            text=black!80,
        }]
        coordinates {   (MNIST-DVS,87.75) 
                        (N-Caltech101,67.50)
                        (CIFAR10-DVS,55.10)
                        }; 
    \addplot+[
        draw=teal!80!black,
        fill=teal!30,
        nodes near coords,
        every node near coord/.append style={
            text=teal!80!black,
        }]
        coordinates {   (MNIST-DVS,94.53) 
                        (N-Caltech101,69.98)
                        (CIFAR10-DVS,56.75)
                        };    
    
    \legend{Float, 8-bit, FG, GLID, FT}
    \end{axis}
    
    \begin{axis}[
            ylabel=BRAMs,
            at=(plot1.below south west), anchor=above north west,
            symbolic x coords={MNIST-DVS,N-Caltech101,CIFAR10-DVS},
            height=3.5cm,
            width=\linewidth,
            grid=major,
            ybar,
            xtick=data, 
            enlarge x limits=0.2,
            ymin=80, ymax=240,
            bar width=7pt,
            nodes near coords={
              \pgfmathprintnumber[
                 fixed zerofill,
                 precision=1
              ]{\pgfplotspointmeta}
            },
            nodes near coords align={vertical}, 
            every node near coord/.append style={
              font=\tiny,
              rotate=90,
              anchor=west,
              xshift=-23pt
            },
            ]

    \addplot+[
        draw=red!80!black,
        fill=red!30,
        nodes near coords,
        every node near coord/.append style={
            text=red!80!black,
        }]
        coordinates {   (MNIST-DVS,223.5) 
                        (N-Caltech101,232.5)
                        (CIFAR10-DVS,229.5)
                        };               
    \addplot+[
        draw=brown!80!black,
        fill=brown!30,
        nodes near coords,
        every node near coord/.append style={
            text=brown!80!black,
        }]
        coordinates {   (MNIST-DVS,165.0) 
                        (N-Caltech101,177)
                        (CIFAR10-DVS,171)
                        };                   
    \addplot+[
        draw=black!80,
        fill=black!30,
        nodes near coords,
        every node near coord/.append style={
            text=black!80,
        }]
        coordinates {   (MNIST-DVS,151.5) 
                        (N-Caltech101,171)
                        (CIFAR10-DVS,163.5)
                        };                
    \addplot+[
        draw=teal!80!black,
        fill=teal!30,
        nodes near coords,
        every node near coord/.append style={
            text=teal!80!black,
        }]
        coordinates {   (MNIST-DVS,151.5) 
                        (N-Caltech101,171)
                        (CIFAR10-DVS,163.5)
                        }; 
    
    \end{axis}
    
    \end{tikzpicture}
    \end{subfigure}
    \hspace{1cm}
    \begin{subfigure}[t]{0.49\textwidth}
    \centering
    \resizebox{\textwidth}{!}{%
      \begin{tabular}{@{}l|cccccc@{}}
        \toprule
        \multirow{2}{*}{Stage} & \multicolumn{2}{c}{CIFAR10-DVS}    & \multicolumn{2}{c}{MNIST-DVS}      & \multicolumn{2}{c}{N-Caltech101} \\
         & Accuracy & BRAMs & Accuracy & BRAMs & Accuracy & BRAMs \\ 
        \midrule
        Float Model   & 0.5840 & \multicolumn{1}{c|}{--}   & 0.9808 & \multicolumn{1}{c|}{--}   & 0.7514 & --    \\
        8-bit Model   & 0.5870 & \multicolumn{1}{c|}{229.5}& 0.9805 & \multicolumn{1}{c|}{223.5}& 0.7514 & 232.5 \\
        FG            & 0.5695 & \multicolumn{1}{c|}{171.0}& 0.9697 & \multicolumn{1}{c|}{165.0}& 0.7107 & 177.0 \\
        GLID          & 0.5510 & \multicolumn{1}{c|}{163.5}& 0.8775 & \multicolumn{1}{c|}{151.5}& 0.6750 & 171.0 \\
        Finetuning    & 0.5675 & \multicolumn{1}{c|}{163.5}& 0.9453 & \multicolumn{1}{c|}{151.5}& 0.6998 & 171.0 \\
        \bottomrule
      \end{tabular}
    }
  \end{subfigure}%

\caption{Prediction accuracy and BRAM usage for hardware implementation at each stage of the proposed pruning method. (Float – baseline Float32 model; 8-bit – quantized model without pruning; FG – Fine Grid Search; GLID – Greedy Layer-wise Iterative Deepening; FT – Fine-Tuning).}
\label{fig:all-stages}
\end{figure}

The final models obtained using the GLID method were subjected to a fine-tuning process to improve accuracy following compression. 
This process involved an additional 10 epochs of training with a low learning rate ($\mathrm{learning\ rate} = 10^{-5}$). 
The final model results after fine-tuning, compared with the results from earlier stages, are presented in Figure~\ref{fig:all-stages}.

The summary of results indicates that for the CIFAR-10 dataset, BRAM resource utilisation was reduced by 28.8\% with only a 1.65\% decrease in accuracy compared to the floating-point model. 
For the MNIST dataset, BRAM reduction reached 31.4\%, with an accuracy drop of 3.55\%. 
For the N-Caltech101 dataset, a 26.5\% saving in BRAM utilisation was achieved at the cost of a 5.18\% reduction in accuracy.

It should be emphasised, however, that the proposed method allows for the selection of an appropriate configuration—both after the first and the second search stages—depending on the specific requirements of the application and the availability of resources on the target platform.

%Zestawienie wyników wskazuje, że dla zbioru CIFAR-10 udało się zmniejszyć zużycie zasobów BRAM o 28.8\% przy jedynie 1.65\% spadku dokładności względem modelu zmiennoprzecinkowego. W przypadku zbioru MNIST redukcja BRAM wyniosła 31.4\%, natomiast dokładność obniżyła się o 3.55\%. Dla zbioru N-Caltech101 odnotowano 26.5\% oszczędności w zużyciu BRAM kosztem 5.18\% niższej dokładności. 

\begin{table}[!t]
\centering
\caption{Best configuration after Greedy Layer-wise Iterative Deepening Search for each dataset.}
\resizebox{0.45\textwidth}{!}{%
\begin{tabular}{@{}lcccccc@{}}
\toprule
      & \multicolumn{2}{c}{MNIST-DVS} & \multicolumn{2}{c}{Cifar10-DVS} & \multicolumn{2}{c}{N-Caltech101} \\ \midrule
Layer & Bits        & Channels        & Bits         & Channels         & Bits          & Channels         \\ \midrule
Conv 1 & 6 & 18  & 6 & 18  & 8 & 24  \\
Conv 2 & 6 & 36  & 6 & 36  & 6 & 33  \\
Conv 3 & 6 & 60  & 6 & 66  & 6 & 72  \\
Conv 4 & 6 & 42  & 6 & 72  & 6 & 57  \\
Conv 5 & 6 & 132 & 6 & 120 & 6 & 132 \\ \bottomrule
\end{tabular}%
}

\label{tab:best-config}
\end{table}

\subsection{Hardware implementation}

To validate the proposed pruning strategy, we developed hardware modules implementing the EFGCN model for the MNIST-DVS classification — both in the baseline form (see Section~\ref{sec:sec:setup}) and in its quantised variant. 
Both configurations were verified for consistency with the software model and subsequently implemented on the heterogeneous ZCU104 platform for 200MHz clock with no timing violation. 
The resource utilisation of both hardware modules is presented in Table~\ref{tab:util} with their overall reduction highlighted.
It should be noted that the total memory resource utilisation in the system is influenced not only by the BRAM blocks used to store the \textit{feature memory} — although these constitute the majority — but also by the resources required for graph representation generation and model weights (which are also reduced). 

The proposed method enabled a BRAM resource reduction of 30.6\% for the entire EFGCN model, with an accuracy drop of 3.55\%. Additionally, due to the decreased dimensions of vectors for multiplication (realised with LUT) and implementation of certain computations with 6-bit precision instead of 8-bit, the utilisation of logic resources (LUTs and flip-flops) was also reduced.
It should be noted, however, that the proposed method results in an increased utilisation of DSP multipliers, as some of the 6-bit multiplications were automatically mapped to these resources during synthesis.

\begin{table}[!t]
\centering
\caption{Utilisation of the baseline and final models implemented for ZCU104 with highlighted key resources reduction.}
\label{tab:util}
\resizebox{0.35\textwidth}{!}{%
\begin{tabular}{@{}l|cccc@{}}
\toprule
Configuration  & LUT & FF & BRAM & DSP \\
\midrule
Baseline model  & 67778 & 23700 & 256 & 96 \\
Final model  & 44468 & 16840 & 177.5 & 148 \\
\midrule
Overall reduction  & 34.4\% & 28.9\% & 30.6\% & -54\% \\
\bottomrule
\end{tabular}%
}
\end{table}

\section{Discussion}
\label{sec:discussion}

The use of the GLID search method alone often overlooks better solutions resulting from further optimisation and rapidly (after only a few steps) reaches the stopping criterion. 
Starting with the FG method allows for a greater initial improvement in performance and the identification of a better search path.  
%We plan further work on design space exploration methods to achieve an optimal solutions. 
We plan further work on design space exploration methods that enable efficient search and facilitate the discovery of even more optimised models.
So far, we have found that approaches based on process-based Bayesian optimisation and Nondominated Sorting Genetic Algorithms (NSGA II and III) did not yield convergent results.

%Dotychczas ustaliliśmy, że metody bazujące na process-based Bayesian optimisation oraz Nondominated Sorting Genetic Algorithms (NSGA II and III) nie zwracały zbierznych rezultatów.

\section{Summary}
\label{sec:summary}

In this work, we present the first exploration of pruning graph neural networks for event-based vision with a focus on hardware architecture and the constraints of the target platform — a heterogeneous FPGA system. Our method integrates structured pruning with quantisation in a way that efficiently utilises the on-chip BRAM and URAM memory resources.

We proposed two design space exploration strategies—\textit{Fine Grid Search} and \textit{Greedy Layer-wise Iterative Deepening Search}—which consider both model accuracy and FPGA resource utilisation. These approaches enable customisation of the network architecture to meet the requirements of specific applications and hardware constraints.

Our strategy was evaluated across different graph input sizes and several popular event-based classification datasets. The results show a BRAM usage reduction of 26.5\% to 31.4\%, with a modest drop in classification accuracy (1.65\% - 5.18\%).

Furthermore, we developed a proof-of-concept hardware module that was verified through simulation and implemented on the ZCU104 platform, achieving an 30.6\% reduction in overall BRAM utilisation, with only a 3.55\% accuracy drop for MNIST-DVS classification.

Future work will focus on advancing hardware-aware pruning and quantisation strategies for graph convolutional networks in event-based vision. Planned extensions include more efficient design space exploration techniques and pruning criteria that consider not only parameter magnitude but also activations, edge characteristics, and spatio-temporal location of inputs.

\bibliographystyle{IEEEtran}
\bibliography{IEEEexample}
% \begin{thebibliography}{00}
% \bibitem{DVS-suvery} Gallego, G., Delbrück, T., Orchard, G., Bartolozzi, C., Taba, B., Censi, A. \& Scaramuzza, D. (2020). Event-based vision: A survey. IEEE transactions on pattern analysis and machine intelligence, 44, 154-180.
% \bibitem{jeziorek_graph} Jeziorek, K., Wzorek, P., Blachut, K., Pinna, A., \& Kryjak, T. (2024). Embedded Graph Convolutional Networks for Real-Time Event Data Processing on SoC FPGAs. arXiv preprint arXiv:2406.07318.
% \bibitem{DVS-GNN} Schaefer, S., Gehrig, D., \& Scaramuzza, D. (2022). Aegnn: Asynchronous event-based graph neural networks. In Proceedings of the IEEE/CVF conference on computer vision and pattern recognition (pp. 12371-12381).
% \bibitem{DVS-FPGA} Kryjak, T. (2024). Event-based vision on FPGAs--a survey. arXiv preprint arXiv:2407.08356.

% \end{thebibliography}

\end{document}